\documentclass{article}
\usepackage[final]{corl_2018} 
\usepackage{microtype}
\usepackage{float}
\usepackage[pdftex]{graphicx}
\usepackage{wrapfig,booktabs}
\usepackage[small]{caption}
\usepackage{xcolor}

\usepackage{setspace}
\usepackage[compact]{titlesec}
\titlespacing{\paragraph}{%
  0pt}{
  0.2\baselineskip}{
  .5em}
\titlespacing{\section}{%
  0pt}{
  0.2\baselineskip}{
  .5em}
\titlespacing{\subsection}{%
  -1pt}{
  0.2\baselineskip}{
  .5em}

\usepackage{parskip}

\usepackage{pgfplots}
\usepgfplotslibrary{fillbetween}

\usepackage{amsmath,amsthm,amssymb}
\usepackage{mathtools}
\usepackage{tabularx,array,threeparttable}
\usepackage{multirow}
\usepackage{hyperref}

\graphicspath{./figures}
\DeclareGraphicsExtensions{.pdf,.jpeg,.png}

\DeclarePairedDelimiter\norm{\lVert}{\rVert}%

\newtheorem{theorem}{Theorem}[section]

\usepackage{algorithm,algpseudocode}
\usepackage{enumitem}

\newcommand{\ww}{\boldsymbol{w}}

\newcommand{\ignore}[1]{}

\usepackage{soul}

\title{Batch Active Preference-Based Learning \\ of Reward Functions}
\author{
  Erdem B\i y\i k\\
  Electrical Engineering\\
  Stanford University\\
  \texttt{ebiyik@stanford.edu}\\
  \And
  Dorsa Sadigh\\
  Computer Science \& Electrical Engineering\\
  Stanford University\\
  \texttt{dorsa@cs.stanford.edu}\\
  }

\begin{document}
\maketitle

\vspace*{-16pt}
\begin{abstract}
Data generation and labeling are usually an expensive part of learning for robotics. While active learning methods are commonly used to tackle the former problem, preference-based learning is a concept that attempts to solve the latter by querying users with preference questions. In this paper, we will develop a new algorithm, \emph{batch active preference-based learning}, that enables efficient learning of reward functions using as few data samples as possible while still having short query generation times. We introduce several approximations to the batch active learning problem, and provide theoretical guarantees for the convergence of our algorithms. Finally, we present our experimental results for a variety of robotics tasks in simulation. Our results suggest that our batch active learning algorithm requires only a few queries that are computed in a short amount of time. We then showcase our algorithm in a study to learn human users' preferences. 

\end{abstract}

\section{Introduction}
\vspace*{-6pt}
	Machine learning algorithms have been quite successful in the past decade. A significant part of this success can be associated to the availability of large amounts of labeled data.
	However, collecting and labeling data can be costly and time-consuming in many fields such as speech recognition \citep{varadarajan2009maximizing}, dialog control \citep{sugiyama2012preference}, text classification \citep{cuong2013active}, image recognition \citep{sener2017geometric}, influence maximization in social networks \citep{chen2013near}, as well as in robotics \citep{sadigh2017active, akrour2012april, jain2015learning}.
	In addition to lack of labeled data, \emph{robot learning} has a few other challenges that makes it particularly difficult. First, humans cannot (and do not) reliably assign a \textit{success value} (reward) to a given robot action. 	
	Furthermore, we cannot simply fall back on collecting demonstrations from humans to learn the desired behavior of a robot since human experts usually provide suboptimal demonstrations or have difficulty operating a robot with more than a few degrees of freedom \citep{akgun2012keyframe,basu2017you}.
	Instead, we use preference-based learning methods that enable us to learn a regression model by using the preferences of users \citep{de2009preference} as opposed to expert demonstrations.
	
	To address the lack of data in robotics applications, we leverage \emph{active preference-based learning} techniques, where we learn from the most informative data to recover humans' preferences of how a robot should act. However, this can be challenging due to the time-inefficiency of most of the active-preference based learning methods.
	The states and actions in every trajectory that is shown to the human naturally are drawn from a continuous space. Previous work has focused on actively synthesizing comparison queries directly from the continuous space~\cite{sadigh2017active}, but these active methods can be quite inefficient. Similary, using the variance of reward estimates to select queries has been explored, but the use of deep reinforcement learning can increase the number of queries required~\cite{christiano2017deep}.
	
Ideally, we would like to develop an algorithm that requires only a few number of queries while generating each query efficiently. 
\begin{quote}
Our insight is that there is a direct tradeoff between \emph{the required number of queries} and the \emph{time it takes to generate each query}.	
\end{quote}
Leveraging this insight, we propose a new algorithm--\emph{batch active preference-based learning}--that balances between the number of queries it requires to learn humans' preferences and the time it spends on generation of each comparison query.
We will actively generate each batch based on the labeled data collected so far. 
Therefore, in our framework, we synthesize and query $b$ pairs of samples, to be compared by the user, at once. In addition, if we are not interested in personalized data collection, the batch query process can be parallelized leading to more efficient results.
Our work differs from the existing batch active learning studies as it involves actively learning a reward function for dynamical systems. Moreover, as we have a continuous set for control inputs and do not have a prior likelihood information of those inputs, we cannot use the \textit{representativeness} measure \citep{wei2015submodularity,elhamifar2016dissimilarity, yang2018single}, which can significantly simplify the problem by reducing it to a submodular optimization.
We summarize our contributions as:
\begin{enumerate}
\item Designing a set of approximation algorithms for efficient batch active learning to learn about humans' preferences from comparison queries.
\item Formalizing the tradeoff between query generation time and the number of queries, and providing convergence guarantees.
\item Experimenting and comparing approximation methods for batch active learning in complex preference based learning tasks.
\item Showcasing our algorithm in predicting human users' preferences in autonomous driving and tossing a ball towards a target. 	
\end{enumerate}

\section{Problem Statement}
\label{sec:citations}
\vspace*{-6pt}
\paragraph{Modeling Choices.}
We start by modeling human preferences about how a robot should act in interaction with other agents. We model these preferences over the actions of an agent in a fully observable dynamical system $\mathcal{D}$. Let $f_{\mathcal{D}}$ denote the dynamics of the system that includes one or multiple robots. Then, $x^{t+1} = f_{\mathcal{D}}(x^t, u_H^t, u_R^t)$, where $u_H^t$ denotes the actions taken by the human, and $u_R^t$ corresponds to the actions of other robots present in the environment.
The state of the system $x^t$ evolves through the dynamics and the actions.

A finite trajectory $\xi\!\in\!\Xi$ is a sequence of continuous state and action pairs $(x^0,\!u_H^0,\!u_R^0)\cdots(x^T,\!u_H^T,\!u_R^T)$ over a finite horizon time $t=0,1,\dots,T$. 
Here $\Xi$ is the set of feasible trajectories, i.e., trajectories that satisfy the dynamics of the system.

\begin{figure}[H]
		\centering
		\vspace*{-10pt}
		\includegraphics[width=0.85\textwidth]{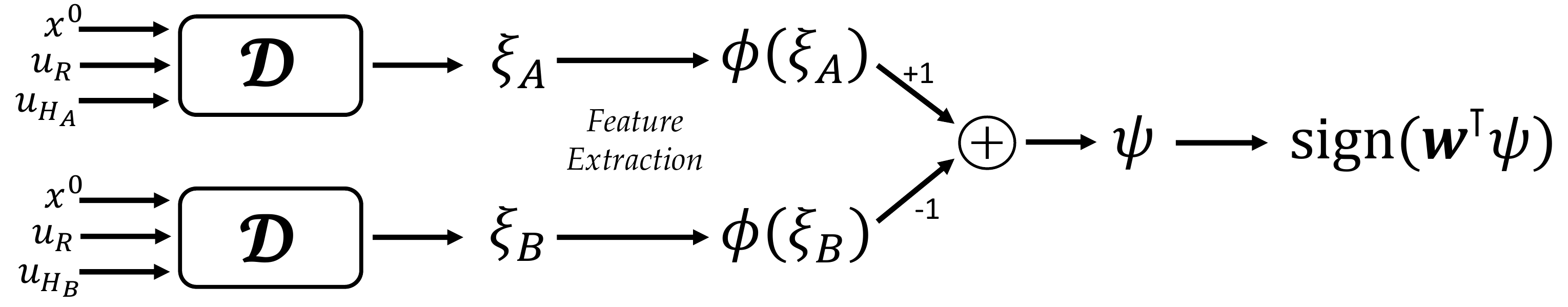}
		\caption{The schematic of the preferences based-learning problem starting from two sample inputs $(x^0, u_{H_A}, u_R)$ and $(x^0, u_{H_B}, u_R)$}
		\label{fig:system}
	\end{figure}

\paragraph{Preference Reward Function.} We model human preferences through a preference reward function $R_H : \Xi \mapsto \mathbb{R}$ that maps a feasible trajectory to a real number corresponding to a score for preference of the input trajectory.
We assume the reward function is a linear combination of a set of features over trajectories $\phi(\xi)$, where $R_H(\xi)\!=\!\ww^\intercal\phi(\xi)$.
	The goal of \emph{preference-based learning} is to learn $R_H(\xi)$, or equivalently $\ww$ through preference queries from a human expert.
	For any $\xi_A$ and $\xi_B$, $R_H(\xi_A)\!>\!R_H(\xi_B)$ if and only if the expert prefers $\xi_A$ over $\xi_B$.
	From this preference encoded as a strict inequality, we can equivalently conclude $\ww^\intercal(\phi(\xi_A)\!-\!\phi(\xi_B))\!>\!0$. We use $\psi$ to refer to this difference: $\psi (\xi_A, \xi_B)\!=\!\phi(\xi_A)\!-\!\phi(\xi_B)$. Therefore, the sign of $\ww^\intercal\psi$ is sufficient to reveal the preference of the expert for every $\xi_A$ and $\xi_B$. We thus let $I\!=\!\mathrm{sign}(\ww^\intercal\psi)$ denote human's input to a query: ``Do you prefer $\xi_A$ over $\xi_B$?". Figure~\ref{fig:system} summarizes the flow that leads to the preference $I$.

	In addition, the input from the human can be noisy due to the uncertainty of her preferences~\citep{sadigh2017active,christiano2017deep,holladay2016active}. A common noise model assumes human's preferences are probabilistic and can be modeled using a softmax function:
	\setlength{\abovedisplayskip}{1pt}
	\setlength{\belowdisplayskip}{1pt}
	\begin{align}
	P(I_i | \ww)
	= \frac1{1+\exp(-I_i\ww^\intercal\psi)}
	\label{eq:human_noise}
	\end{align}
	where $I_i\!=\!\mathrm{sign}(\boldsymbol{w}^T\psi_i)$ represents human's preference on the $i^{th}$ query with trajectories $\xi_A$ and $\xi_B$.

\paragraph{Approach Overview.} 
In many robotics tasks, we are interested in learning a model of the humans' preferences about the robots' trajectories. This model can  be learned through inverse reinforcement learning (IRL), where a reward function $R_H$ is learned directly from the human demonstrating how to operate a robot~\cite{ziebart2008maximum,levine2012continuous,sadigh2016planning-journal,Sadigh:EECS-2017-143}. However, learning a reward function from humans' preferences as opposed to demonstrations can be more favorable for a few reasons. First, providing demonstrations for robots with higher degrees of freedom can be quite challenging even for human experts~\cite{akgun2012keyframe}. Furthermore, humans' preferences tend to defer from their demonstrations~\cite{basu2017hri}.

We plan to leverage \emph{active} preference-based techniques to synthesize pairwise queries over the continuous space of trajectories for the goal of efficiently learning humans' preferences~\cite{sadigh2017active,akrour2011preference,furnkranz2012preference,sugiyama2012preference,wilson2012bayesian}.
However, there is a tradeoff between the time spent to generate a query and the number of queries required until converging to the human's preference reward function. 
Although actively synthesizing queries can reduce the total number of queries, generating each query can be quite time-consuming, which can make the approach impractical by creating a slow interaction with humans.

Instead, we propose a time-efficient method, \emph{batch} active learning, that balances between minimizing the number of queries and being time-efficient in its interaction with the human expert.
Batch active learning has two main benefits: i) Creating a batch of queries can create a more \emph{time-efficient} interaction with the human. ii)  The procedure can be \emph{parallelized} when we look for the preferences of a population of humans.
 
\section{Time-Efficient Active Learning for Synthesizing Queries}
\paragraph{Actively Synthesizing Pairwise Queries.}
In active preference-based learning, the goal is to synthesize the next pairwise query to ask a human expert to maximize the information received. 
While optimal querying is NP-hard \cite{ailon2012active}, there exist techniques that pose the problem as a submodular optimization, where suboptimal solutions that work well in practice exist. 
We follow the work in~\cite{sadigh2017active}, where we model active preference-based learning as a \emph{maximum volume removal} problem. 

The goal is to search for the human's preference reward function $R_H = \ww^ \intercal \phi (\xi)$ by actively querying the human. We  let $p(\ww)$ be the distribution of the unknown weight vector $\ww$. Since $\ww$ and $c\ww$ yield to the same preferences for a positive constant $c$, we constrain the prior such that $\norm{\ww}_2\leq1$. Every query provides a human input $I_i$, which then enables us to perform a Bayesian update of this distribution as $p(\ww|I_i) \propto p(I_i|\ww)p(\ww)$. Since we do not know the shape of $p(\ww)$ and cannot differentiate through it, we sample $M$ values from $p(\ww)$ using an adaptive Metropolis algorithm \citep{haario2001adaptive}. In order to speed up this sampling process, we approximate $p(I_i | \ww)$ as $\min(1,\exp(I_i\ww^\intercal\psi))$.
Generating the next most informative query can  be formulated as maximizing the minimum volume removed from the distribution of $\ww$ at every step. We note that every query, i.e., a pair of trajectories $(\xi_A,\xi_B)$ is parameterized by the initial state $x^0$, a set of actions for all the other agents $u_R$, and the two sequence of actions $u_{H_A}$ and $u_{H_B}$ corresponding to $\xi_A$ and $\xi_B$ respectively.
The query selection problem in the $i^{th}$ iteration can then be formulated as:
	\begin{align}
	\max_{x^0,u_R,u_{H_A},u_{H_B}}\:\min\{\mathbb{E}[1-p(I_i|\ww)], \mathbb{E}[1-p(-I_i|\ww)]\}
	\label{eq:optimization}
	\end{align}
	with an appropriate feasibility constraint. Here the inner optimization (minimum between two volumes for the two choices of user input) provides robustness against the user's preference on the query, the outer optimization ensures the maximum volume removal, where \textit{volume} refers to the unnormalized distribution $p(\ww)$. This sample selection approach is based on the expected value of information of the query \citep{krueger2016active} and the optimization can be solved using a Quasi-Newton method~\citep{andrew2007scalable}.
\paragraph{Batch Active Learning.}
Actively generating a new query requires solving the optimization in equation~\eqref{eq:optimization} and running the adaptive Metropolis algorithm for sampling. Performing these operations for every single query synthesis can be quite slow and not very practical while interacting with a human expert in real-time. The human has to wait for the solution of optimization before being able to respond to the next query.  
 Our insight is that we can in fact balance between the number of queries required for convergence to $R_H$ and the time required to generate each query.
We construct this balance by introducing a \emph{batch active learning} approach, where $b$ queries are simultaneously synthesized at a time. The batch approach can significantly reduce the total time required for the satisfactory estimation of $\ww$ at the expense of increasing the number of queries.


Since small perturbations of the inputs could lead to only minor changes in the objective of equation~\eqref{eq:optimization}, continuous optimization of this objective can result in generating same or sufficiently similar queries within a batch. We thus fall back to a discretization method.  We discretize the space of trajectories by sampling $K$ pairs of trajectories from the input space of $\xi = (x^0, u_H, u_R)$. While increasing $K$ yields more accurate optimization results, computation time increases linearly with $K$.
	
A similar viewpoint to optimization in~\eqref{eq:optimization} is to use the notion of information entropy. As in uncertainty sampling, a similar interpretation of equation~\eqref{eq:optimization} is to find a set of feasible queries that maximize the conditional entropy $\mathcal{H}(I_i|\ww)$. Following this conditional entropy framework, we formalize the batch active learning problem as the solution of the following optimization:	
	\begin{align}
	\max_{\xi_{{ib+1}_A},\xi_{{ib+1}_B},\dots,\xi_{{(i+1)b}_A},\xi_{{(i+1)b}_B}} \mathcal{H}(I_{ib+1}, I_{ib+2}, \dots, I_{(i+1)b} | \boldsymbol{w})
	\label{eq:optimization_batch}
	\end{align} 
	for the $(i+1)^{th}$ batch with the appropriate feasibility constraint. This problem is known to be computationally hard \citep{cuong2013active,chen2013near} \textemdash it requires an exhaustive search which is intractable in practice, since the search space is exponentially large \citep{guo2008discriminative}.

\subsection{Algorithms for Time-Efficient Batch Active Learning}
\begin{figure}
	\centering
	\includegraphics[width=\textwidth]{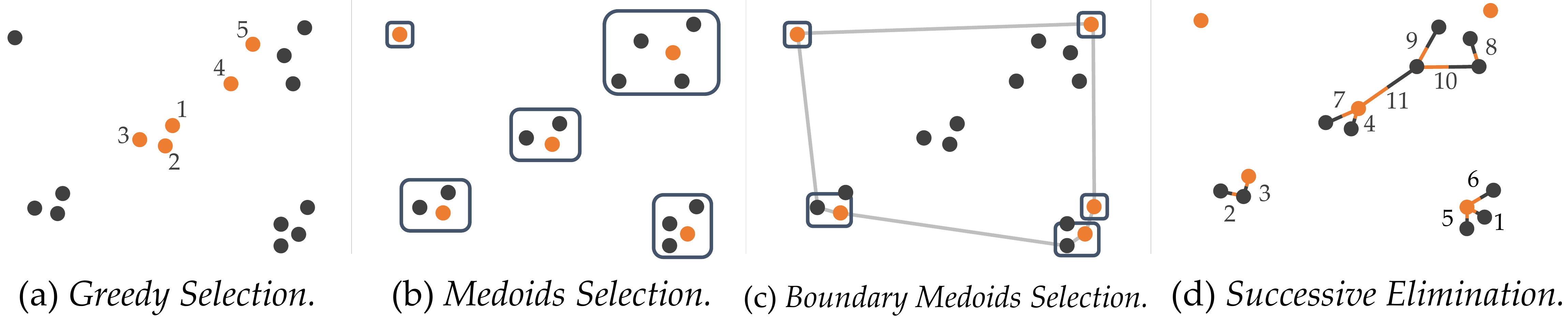}
	\vspace*{-16pt}
	\caption{
	Visualizations of the selection process of batch active learning. A simple 2D space with 16 different $\psi$ values that correspond to inputs individually maximizing the conditional entropy. The goal is to select a batch of $b\!=\!5$ that will near-optimally maximize the joint conditional entropy. The selected samples are shown in orange. 
	(a) Greedy Selection, (b) Medoids Selection, (c) Boundary Medoids Selection, (d) Successive Elimination.
	}
	\label{fig:method_visuals}
	\vspace*{-8pt}
\end{figure}

\vspace*{-6pt}
We now describe a set of methods in increasing order of complexity to provide an approximation to the batch active learning problem. Figure~\ref{fig:method_visuals} visualizes each approach for a small set of samples.

\paragraph{Greedy Selection.}
The simplest method to approximate the batch learning problem in equation~\eqref{eq:optimization_batch} is using a greedy strategy. In the greedy selection approach, we conveniently assume the $b$ different human inputs are independent of each other. Of course this is not a valid assumption, but the independence assumption creates the following approximation, where we need to choose the $b$-many maximizers of equation~\eqref{eq:optimization} among the $K$ samples:
\begin{align} 
\max_{\xi_{{ib+1}_A},\xi_{{ib+1}_B}} \mathcal{H}(I_{ib+1}|\ww) + \dots + \max_{\xi_{{(i+1)b}_A},\xi_{{(i+1)b}_B}} \mathcal{H}(I_{(i+1)b}|\ww)
\label{eq:optimization_batch_greedy}
\end{align}
with an additional set of constraints that specify the trajectory sets $(\xi_A,\xi_B)$ are different among queries. 
While this method can easily be employed, it is suboptimal as redundant samples can be selected together in the same batch, since these similar queries are likely to lead to high entropy values. For instance, as shown in Fig.~\ref{fig:method_visuals}~(a) the $5$ orange samples chosen are all going to be close to the center where there is high conditional entropy.

\paragraph{Medoid Selection.}
	To avoid the redundancy in the samples created by the greedy selection, we need to increase the dissimilarity between the selected batch samples. Our insight is to define a new approach, \emph{Medoid Selection}, that leverages clustering as a similarity measure between the samples.  In this method, we let $\mathcal{G}_B$ be the set of $\psi$-vectors that correspond to $B$ samples selected using the greedy selection strategy, where $B>b$. With the goal of picking the most dissimilar samples, we cluster $\mathcal{G}_B$ into $b$ clusters, using standard Euclidean distance. We then restrict ourselves to only selecting one element from each cluster, which prevents us from selecting very similar trajectories. One can think of using the well-known K-means algorithm \citep{lloyd1982least} for clustering and then selecting the centroid of each cluster. However, these centroids are not necessarily from the set of greedily selected samples, so they can have lower expected information.
	

	Instead, we use the K-medoids algorithm \citep{kaufman1987clustering, bauckhage2015numpy} which again clusters the samples into $b$ sets. The main difference between K-means and K-medoids is that K-medoids enables us to select medoids as opposed to the centroids, which are points \textit{in the set} $\mathcal{G}_B$ that minimize the average distance to the other points in the same cluster.
	Fig.~\ref{fig:method_visuals}~(b) shows the medoid selection approach, where $5$ orange points are selected from $5$ clusters.

\paragraph{Boundary Medoid Selection.}
	We note that picking the medoid of each cluster is not the best option for increasing dissimilarity \textemdash instead, we can further exploit clustering to select samples more effectively. In the \emph{Boundary Medoid Selection} method, we propose restricting the selection to be only from the boundary of the convex hull of $\mathcal{G}_B$.
	This selection criteria can separate out the sample points from each other on average.
	We note that when $\dim(\psi)$ is large enough, most of the clusters will have points on the boundary. We thus propose the following modifications to the medoid selection algorithm.
 	The first step is to only select the points that are on the boundary of the convex hull of $\mathcal{G}_B$. We then apply K-medoids with $b$ clusters over the points on the boundary and finally only accept the cluster medoids as the selected samples.
As shown in Fig.~\ref{fig:method_visuals}~(c), we first find $b=5$ clusters over the points on the boundary of the convex hull of $\mathcal{G}_B$. We then select the medoid of those $5$ clusters created over the boundary points.
	
\paragraph{Successive Elimination.}
The main goal of batch active learning as described in the previous methods is to select $b$ points that will maximize the average distance among them out of the $B$ samples in $\mathcal{G}_B$. This problem is called \textit{max-sum diversification} in literature, known to be NP-hard \citep{gollapudi2009axiomatic, borodin2012max}.
	
	What makes our batch active learning problem special and different from standard max-sum diversification is that we can compute the conditional entropy $\mathcal{H}(I_i|\ww)$ for each potential pair of trajectories, which corresponds to $\psi_i$.
	The conditional entropy is a metric that models how much a query is preferred to be in the final batch.
	We propose a novel method that leverages the conditional entropy to successively eliminate samples for the goal of obtaining a satisfactory diversified set. We refer to this algorithm as \emph{Successive Elimination}. At every iteration of the algorithm, we select two closest points in $\mathcal{G}_B$, and remove the one with lower information entropy. We repeat this procedure until $b$ points are left in the set resulting in the $b$ samples in our final batch, which efficiently increases the diversity among queries.	Fig.~\ref{fig:method_visuals}~(d) shows the successive pairwise comparisons between two samples based on their corresponding conditional entropy. In every pairwise comparison, we eliminate the sample shown with gray edge, keeping the point with the orange edge. The numbers show the order of comparisons done before finding $b\!=\!5$ optimally different sample points shown in orange.

\subsection{Convergence Guarantees}
	\begin{theorem}
		Under the following assumptions:
		\begin{enumerate}[leftmargin=0.3cm,itemindent=0.3cm]
			\item The error introduced by the sampling of input space is ignored,
			\item The function that updates the distribution of $\boldsymbol{w}$, and the noise that human inputs have are $p(I_i | \boldsymbol{w})$ as given in Eq.~\eqref{eq:human_noise}; and the error introduced by approximation of noise model is ignored,
			\item The errors introduced by the sampling of $\boldsymbol{w}$'s and non-convex optimization is ignored,
		\end{enumerate}
		greedy selection and successive elimination algorithms remove at least $1\!-\!\epsilon$ times as much volume as removed by the best adaptive strategy after $b\ln(\frac1{\epsilon})$ times as many queries.
	\end{theorem}
	\vspace*{-4pt}
	\begin{proof}
		In greedy selection and successive elimination methods, the conditional entropy maximizer query $(\xi_A^*,\xi_B^*)$ out of $K$ possible queries will always remain in the resulting batch of size $b$, because the queries will be removed only if they have lower entropy than some other queries in the set. By assumption 1, we have $(\xi_A^*,\xi_B^*)$ as the maximizer over the continuous control inputs set. In \citep{sadigh2017active}, it has been proven by using the ideas from submodular function maximization literature \citep{krause2014submodular} that if we make the single query $(\xi_A^*,\xi_B^*)$ at each iteration, at least $1\!-\!\epsilon$ times as much volume as removed by the best adaptive strategy will be removed after $\ln(\frac1{\epsilon})$ times as many iterations. The proof is complete with the pessimistic approach accepting other $b\!-\!1$ queries will remove no volume.
	\end{proof}

\section{Simulations and Experiments}
\vspace*{-6pt}
	\begin{figure}
		\centering
		\includegraphics[width=\textwidth]{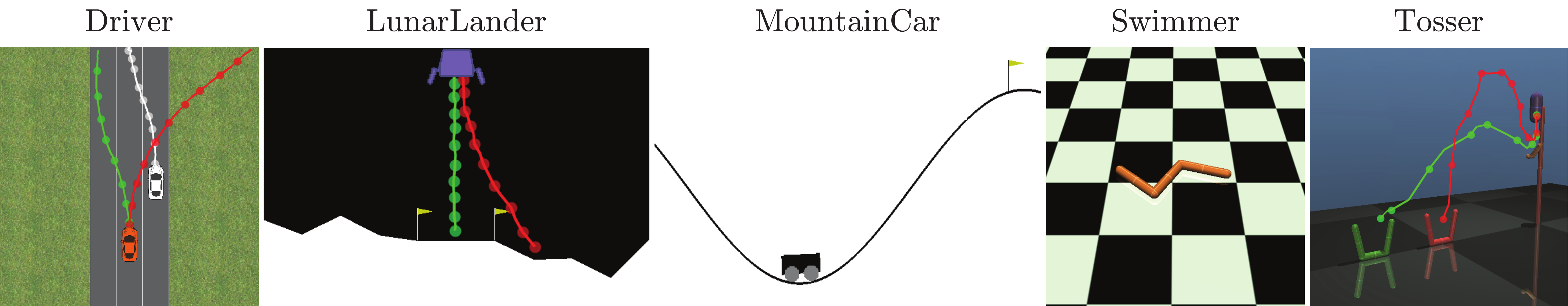}
		\vspace*{-16pt}
		\caption{
		Views from each task. (a) Driver, (b) Lunar Lander, (c) Mountain Car, (d) Swimmer, and (e) Tosser.
			}
		\label{fig:experiment_visuals}
		\vspace*{-10pt}
	\end{figure}

	\paragraph{Experimental Setup.} We have performed several simulations and experiments to compare the methods we propose and to demonstrate their performance. The code is available online\footnote{See  \url{http://github.com/Stanford-ILIAD/batch-active-preference-based-learning}}. In our experiments, we set $b\!=\!10$, $B\!=\!20b$ and $M\!=\!1000$.  We sample the input space with $K\!=\!5\!\times\!10^5$ and compute the corresponding $\psi$ vectors once, and use this sampled set for every experiment and iteration. To acquire more realistic trajectories, we fix $u_R$ when other agents exist in the experiment.

\paragraph{Alignment Metric.} For our simulations, we generate a synthetic random $\ww_{\textrm{true}}$ vector as our true preference vector. We have used the following alignment metric \citep{sadigh2017active} in order to compare non-batch active, batch active and random query learning methods, where all queries are selected uniformly random over all feasible trajectories.
	\begin{align}
	m = \frac{\ww_{\textrm{true}}^\intercal \hat{\ww}}{\norm{\ww_{\textrm{true}}}_2\norm{\hat{\ww}}_2}
	\end{align}
	where $\hat{\ww}$ is $\mathbb{E}[\ww]$ based on the estimate of the learned distribution of $\ww$. We note that this alignment metric can be used to test convergence, because the value of $m$ being close to $1$ means the estimate of $\ww$ is very close to (aligned with) the true weight vector.

\subsection{Tasks}
We perform experiments in different simulation environments. Fig.~\ref{fig:experiment_visuals} visualizes each of the experiments with some sample trajectories. We now briefly describe the environments.

	\paragraph{Linear Dynamical System (LDS).}
	We assess the performance of our methods on an LDS:	
	\begin{align}
	x^{t+1} = Ax^t + Bu^t,\quad y^t = Cx^t + Du^t
	\end{align}
	For a fair comparison between the proposed methods independent of the dynamical system, we want $\phi(\xi)$ to uniformly cover its range when the control inputs are uniformly distributed over their possible values. We thus set $A$, $B$ and $C$ to be zeros matrices and $D$ to be the identity matrix. Then we treat $y^0$ as $\phi(\xi)$. Therefore, the control inputs are equal to the features over trajectories, and optimizing over control inputs or features is equivalent.
	We repeat this simulation $10$ times, and use non-parametric Wilcoxon signed-rank tests over these $10$ simulations and $9$ different $N$ to assess significant differences \citep{wilcoxon1945individual}.

	\vspace*{-3pt}
	\paragraph{Driving Simulator.} We use the 2D driving simulator~\citep{sadigh2016planning}, shown in Fig.~\ref{fig:experiment_visuals}~(a). We use features corresponding to distance to the closest lane, speed, heading angle, and distance to the other vehicles. Two sample trajectories are shown in red and green in Fig.~\ref{fig:experiment_visuals}~(a). In addition, the white trajectory shows the state and actions ($u_R$) of the other vehicle.
	\vspace*{-3pt}
	\paragraph{Lunar Lander.} We use OpenAI Gym's continuous Lunar Lander \citep{brockman2016openai}. We also use features corresponding to final heading angle, final distance to landing pad, total rotation, path length, final vertical speed, and flight duration. Sample trajectories are shown in Fig.~\ref{fig:experiment_visuals}~(b).	\vspace*{-3pt}
	\paragraph{Mountain Car.} We use OpenAI Gym's continuous Mountain Car \citep{brockman2016openai}. The features are maximum range in the positive direction, maximum range in the negative direction, time to reach the flag.
	\vspace*{-3pt}
	\paragraph{Swimmer.} We use OpenAI Gym's Swimmer \citep{brockman2016openai}. Similarly we use features corresponding to horizontal displacement, vertical displacement, total distance traveled.
	\vspace*{-3pt}
	\paragraph{Tosser.} We use MuJoCo's Tosser \citep{todorov2012mujoco}. The features we use are maximum horizontal range, maximum altitude, the sum of angular displacements at each timestep, final distance to closest basket. The two red and green trajectories in Fig.~\ref{fig:experiment_visuals}~(e) correspond to synthesized queries showing different preferences for what basket to toss the ball to.		
	
\subsection{Experiment Results}
\label{sec:result}
	\begin{wrapfigure}{r}{0.50\textwidth}
	\centering
		\vspace*{-44pt}
		\hspace*{-6pt}
		\includegraphics[width=0.52\textwidth]{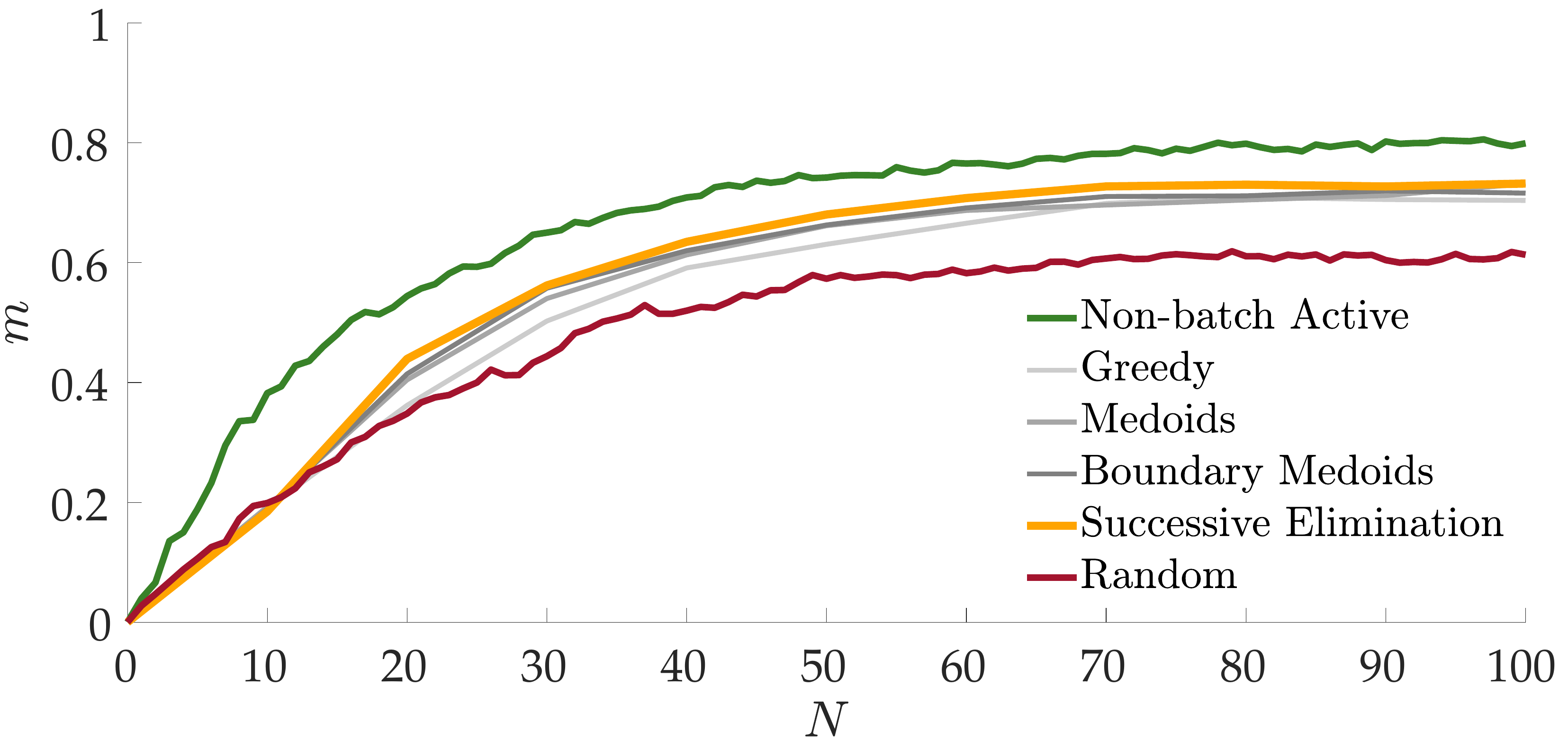}
		\vspace*{-16pt}
		\caption{The performance of each algorithm was averaged over $10$ different runs with LDS. The proposed batch methods perform better than the random querying baseline and worse than the non-batch active methods. 
		}
		\vspace*{-4pt}
		\label{fig:simulation}
	\end{wrapfigure}
	
	For the LDS simulations, we assume human's preference is noisy as discussed in Eq.~\eqref{eq:human_noise}. For other tasks, we assume an oracle user who knows the true weights $\ww_{\textrm{true}}$ and responds to queries with no error.

	Figure~\ref{fig:simulation} shows the number of queries that result in a corresponding alignment value $m$ for each method in the LDS environment. While non-batch active version as described in \citep{sadigh2017active} outperforms all other methods as it performs the optimization for each and every query, successive elimination method seems to improve over the remaining methods on average. The performance of batch-mode active methods are ordered from worst to best as \emph{greedy}, \emph{medoids}, \emph{boundary medoids}, and \emph{successive elimination}. While the last three algorithms are significantly better than greedy method ($p<0.05$), and successive elimination is significantly better than medoid selection ($p<0.05$); the significance tests for other comparisons are somewhat significant ($p<0.13$). This suggests successive elimination increases diversity without sacrificing informative queries.

\begin{figure}
		\centering
		\includegraphics[width=\textwidth]{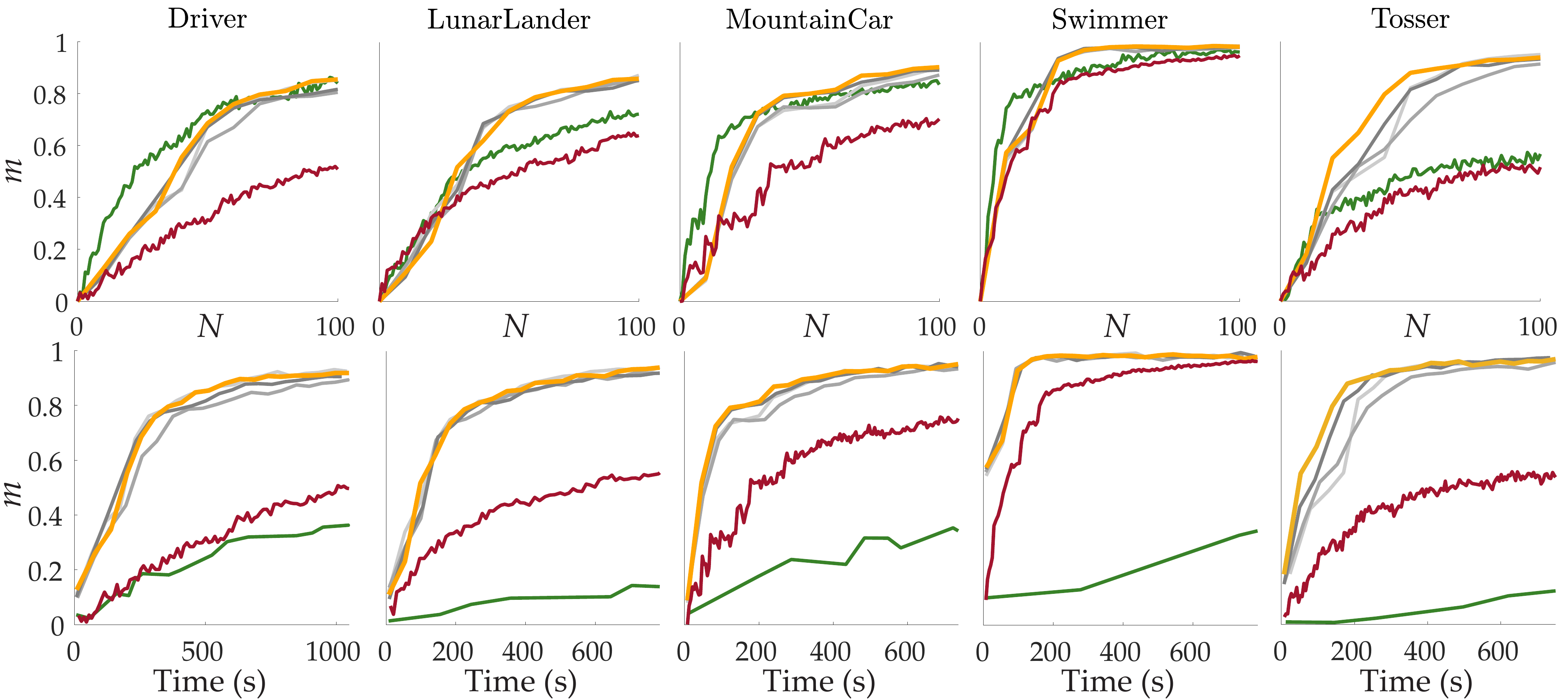}
		\vspace*{-16pt}
		\caption{The performance of each algorithm is shown for all $5$ tasks. This figure uses the same legend as Fig.~\ref{fig:simulation}. \textbf{Top row:} While it is difficult to compare batch active algorithms in the environments other than \textit{MountainCar} and \textit{Tosser}, where successive elimination is superior, we also note non-batch active method performs poorly on \textit{LunarLander} and \textit{Tosser}. \textbf{Bottom Row}: Non-batch active learning method is slow due to the optimization and adaptive metropolis algorithm involved in each iteration, whereas random querying performs poorly due to redundant queries. Batch active methods clearly outperform both of them.}
		\label{fig:task_simulations}
	\end{figure}

\begin{wrapfigure}{r}{0.55\textwidth}
	\centering
	\vspace*{-22pt}
	\hspace*{-6pt}
	\includegraphics[width=0.57\textwidth]{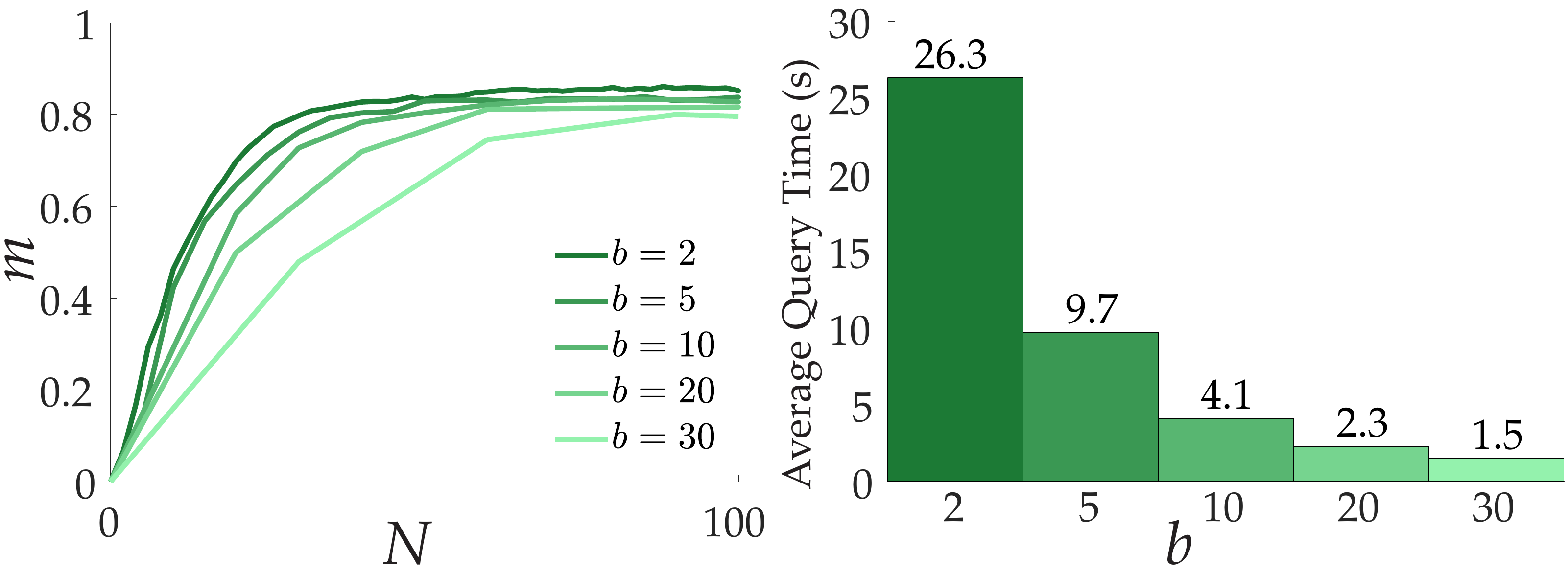}
	\vspace*{-18pt}
	\caption{The performance of successive elimination algorithm with varying $b$ values was averaged over $10$ different runs with LDS.
	}
	\vspace*{-6pt}
	\label{fig:batch_size}
\end{wrapfigure}
We show the results of our experiments in all $5$ environments in Fig.~\ref{fig:task_simulations} and Table~\ref{tab:average_query_times}. Fig.~\ref{fig:task_simulations}~(a) shows the convergence to the true weights $\ww_{\textrm{true}}$ as the number of samples $N$ increases (similar to Fig.~\ref{fig:simulation}). Interestingly, non-batch active learning performs suboptimally in \textit{LunarLander} and \textit{Tosser}.  We believe this can be due to the non-convex optimization involved in non-batch methods leading to suboptimal behavior. The proposed batch active learning methods overcome this issue as they sample from the input space.
	
	\begin{table}
		\centering
		\vspace*{-6pt}
		\caption{Average Query Times (seconds)}
		\label{tab:average_query_times}
		\begin{tabular}{| c | c | c | c | c | c |} 
			\hline
			\multirow{2}{*}{\bfseries Task Name} & \multirow{2}{*}{\bfseries Non-Batch} & \multicolumn{4}{c|}{\bfseries Batch Active Learning}\\
			\cline{3-6}
			& & \emph{Greedy} & \emph{Medoids} & \emph{Boundary Med.} & \emph{Succ. Elimination}\\
			\hline
			\emph{Driver} & 79.2 & 5.4 & 5.7 & 5.3 & 5.5 \\ 
			\hline
			\emph{LunarLander} & 177.4 & 4.1 & 4.1 & 4.2 & 4.1 \\
			\hline
			\emph{MountainCar} & 96.4 & 3.8 & 4.0 & 4.0 & 3.8 \\
			\hline
			\emph{Swimmer} & 188.9 & 3.8 & 3.9 & 4.0 & 4.1 \\
			\hline
			\emph{Tosser} & 149.3 & 4.1 & 4.3 & 3.8 & 3.9 \\
			\hline
		\end{tabular}
		\vspace*{-14pt}
	\end{table}
	
	Figure~\ref{fig:task_simulations}~(b) and Table~\ref{tab:average_query_times} evaluate the computation time required for querying. It is clearly visible from Fig.~\ref{fig:task_simulations}~(b) that batch active learning makes the process much faster than the non-batch active version and random querying. 
	Hence, batch active learning is preferred over other methods as it balances the tradeoff between the number of queries required and the time it takes to compute them. This tradeoff can be seen from Fig.~\ref{fig:batch_size} where we simulated LDS with varying $b$ values.

\subsection{User Preferences}
	\begin{wrapfigure}{r}{0.5\textwidth}
		\centering
		\vspace*{-36pt}
		\includegraphics[width=0.5\textwidth]{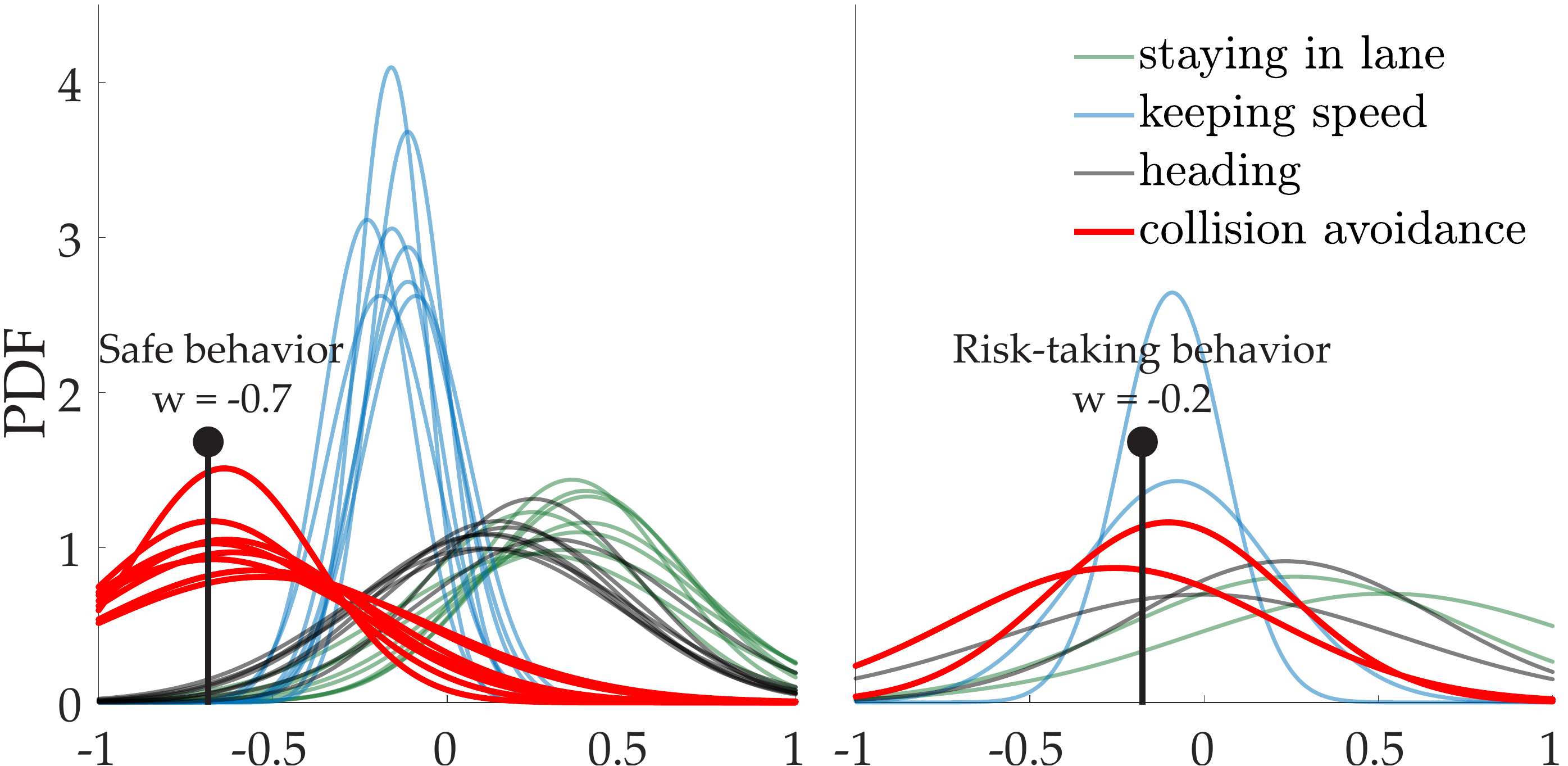}
		\vspace*{-14pt}
		\caption{User preferences on \textit{Driver} task are grouped into two sets. The first set shows the preferences conforming with the natural driving behavior. The second set is comprised of data from two users one of whom preferred collisions over leaving the road and the other regarded some collisions as near-misses and thought they can be acceptable to keep speed. It can be seen that the uncertainty in their learned preferences is higher.}
		\vspace*{-6pt}
		\label{fig:user_study_driver}
	\end{wrapfigure}
In addition to our simulation results using a synthetic $\ww_{\textrm{true}}$, we perform a user study to learn humans' preferences for the  \textit{Driver} and \textit{Tosser} environments. This experiment is mainly designed to show the ability of our framework to learn humans' preferences.

\paragraph{Setup.} We recruited $10$ users who responded to $150$ queries generated by successive elimination algorithm for \textit{Driver} and \textit{Tosser} environments.

\paragraph{Driver Preferences.}
Using successive elimination, we are able to learn humans' driving preferences. We use four features corresponding to the vehicle staying within its lane, having high speed, having a straight heading, and avoiding collisions. Our results show the preferences of users are very close to each other as this task mainly models natural driving behavior. This is consistent with results shown in \citep{sadigh2017active}, where non-batch techniques are used. We noticed a few differences between the driving behaviors as shown in Fig.~\ref{fig:user_study_driver}. This figure shows the distribution of the weights for the four features after $150$ queries. Two of the users (plot on the right) seem to have slightly different preferences about collision avoidance, which can correspond to more aggressive driving style.

We observed $70$ queries were enough for converging to safe and sensible driving in the defined scenario where we fix the speed and let the system optimize steering. The optimized driving with different number of queries can be watched on \url{https://youtu.be/MaswyWRep5g}.

\begin{figure}[tb]
		\centering
		\vspace*{-16pt}
		\includegraphics[width=\textwidth]{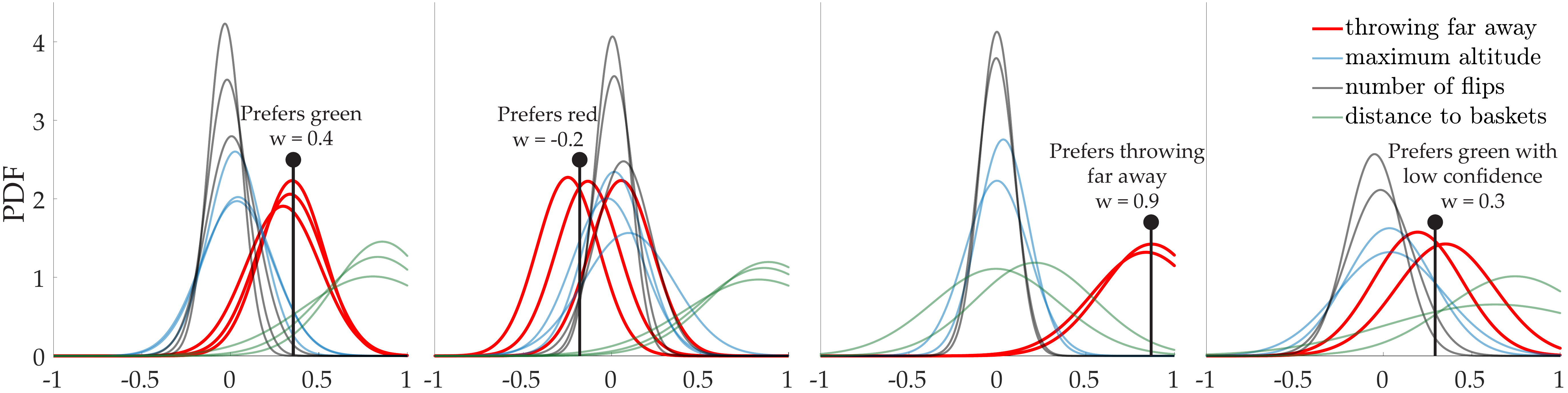}
		\vspace*{-14pt}
		\caption{User preferences on \textit{Tosser} task are grouped into four sets. The first set shows the preferences of people who aimed at throwing the ball into the green basket but accepted throwing into the other basket is better than not throwing into any baskets. The second set is comprised of data from three users who preferred the red basket. In the third group, the users preferred the green basket over the red one, but also accepted throwing far away is better than throwing into the red basket, because it is an attempt for the green basket. The fourth group is similar to the first group; however the confidence over preferences is much less, because the users were not sure about how to compare the cases where the ball was dropped between the baskets in one of the trajectories.}
		\vspace*{-6pt}
		\label{fig:user_study_tosser}
	\end{figure}

\paragraph{Tosser Preferences.} Similarly, we use successive elimination to learn humans' preferences on the tosser task. The four features used correspond to: throwing the ball far away, maximum altitude of the ball, number of flips, and distance to the basket. These features are sufficient to learn interesting tossing preferences as shown in Fig.~\ref{fig:user_study_tosser}.
For demonstration purposes, we optimize the control inputs with respect to the preferences of two of the users, one of whom prefers the green basket while the other prefers the red one (see Fig.~\ref{fig:experiment_visuals}~(e)). We note 100 queries were enough to see reasonable convergence. The evolution of the learning can be watched on \url{https://youtu.be/cQ7vvUg9rU4}.
	
\section{Discussion}
\vspace*{-6pt}
	\paragraph{Summary.} In this work, we have proposed an end-to-end method to efficiently learn humans' preferences on dynamical systems. Compared to the previous studies, our method requires only a small number of queries which are generated in a reasonable amount of time. We provide theoretical guarantees for convergence of our algorithm, and demonstrate its performance in simulation.

\paragraph{Limitations.} 
In our experiments,
we sample the control space in advance for batch-mode active learning methods, while we still employ the optimization formulation for the non-batch active version. It can be argued that this creates a bias on the computational times. However, there are two points that make batch techniques more efficient than the non-batch version. First, this sampling process can be easily parallelized. Second, even if we used predefined samples for non-batch method, it would be still inefficient due to adaptive Metropolis algorithm and discrete optimization running for each query, which cannot be parallelized across queries. It can also be inferred from Fig.~\ref{fig:batch_size} that non-batch active learning with sampling the control space would take a significantly longer running time compared to batch versions. We note the use of sampling would reduce the performance of non-batch active learning, while it is currently the best we can do for batch version.


\paragraph{Future directions.} In this study, we used a fixed batch-size. However, we know the first queries are more informative than the following queries. Therefore, instead of starting with $b$ random queries, one could start with smaller batch sizes and increase over time. This would both make the first queries more informative and the following queries computationally faster.

The algorithms we described in this work can be easily implemented when appropriate simulators are available. For the cases where safety-critical dynamical systems are to be used,  further research is warranted to ensure that the optimization is not evaluated with unsafe inputs.

We also note the procedural similarity between our successive elimination algorithm and Mat{\'e}rn processes \cite{matern2013spatial}, which also points out a potential use for determinantal point processes for diversity within batches \cite{kulesza2012determinantal,zhang2017determinantal}.

Lastly, we used handcrafted feature transformations in this study. In the future we plan to learn those transformations along with preferences, i.e. to learn the reward function directly from trajectories, by developing batch techniques that use as few queries as possible generated in a short amount of time.

\acknowledgments{The authors would like to acknowledge FLI grant RFP2-000. Toyota Research Institute (``TRI")  provided funds to assist the authors with their research but this article solely reflects the opinions and conclusions of its authors and not TRI or any other Toyota entity. Erdem B\i y\i k is partly supported by the Stanford School of Engineering James D. Plummer Graduate Fellowship.}
\newpage
\bibliography{refs,sadigh}  

\end{document}